\documentclass[10pt,twocolumn,letterpaper]{article}

\usepackage{iccv}
\usepackage{times}
\usepackage{epsfig}
\usepackage{graphicx}
\usepackage{amsmath}
\usepackage{amssymb}

\usepackage{booktabs}
\usepackage{multirow}
\usepackage{verbatim}
\usepackage{mathrsfs}
\usepackage{mathtools} % for $\coloneqq$
\usepackage{cite} % to automatically order the citation's index

\DeclareMathOperator*{\argmin}{arg\,min}

\usepackage[breaklinks=true,bookmarks=false,pagebackref,colorlinks]{hyperref}

\usepackage[capitalize]{cleveref}
\crefname{section}{Sec.}{Secs.}
\Crefname{section}{Section}{Sections}
\Crefname{table}{Table}{Tables}
\crefname{table}{Tab.}{Tabs.}

\iccvfinalcopy % *** Uncomment this line for the final submission

\def\iccvPaperID{9947} % *** Enter the ICCV Paper ID here

\ificcvfinal\fi

\begin{document}

%%%%%%%%% TITLE
\title{Privacy-Preserving Face Recognition Using Random Frequency Components} %Randomized Frequency Components

\author{
Yuxi Mi$^{1}$\thanks{Equal contributions.}\quad
Yuge Huang$^{2}\footnotemark[1]$\quad
Jiazhen Ji$^{2}$\quad
Minyi Zhao$^{1}$ \\
Jiaxiang Wu$^{2}$\quad
Xingkun Xu$^{2}$\quad
Shouhong Ding$^{2}$\quad
Shuigeng Zhou$^{1}$\thanks{Corresponding author.}
\\
$^{1}$ Fudan University \quad
$^{2}$ Tencent Youtu Lab
\\
{\tt\small \{yxmi20, zhaomy20, sgzhou\}@fudan.edu.cn} \\
{\tt\small \{yugehuang, royji, willjxwu, xingkunxu, ericshding\}@tencent.com}
}

% \and

\maketitle
\ificcvfinal\thispagestyle{empty}\fi

%%%%%%%%% ABSTRACT
\begin{abstract}

% This paper provides an in-depth study of two privacy goals.

The ubiquitous use of face recognition has sparked increasing privacy concerns, as unauthorized access to sensitive face images could compromise the information of individuals. This paper presents an in-depth study of the privacy protection of face images' visual information and against recovery. Drawing on the perceptual disparity between humans and models, we propose to conceal visual information by pruning human-perceivable low-frequency components. For impeding recovery, we first elucidate the seeming paradox between reducing model-exploitable information and retaining high recognition accuracy. Based on recent theoretical insights and our observation on model attention, we propose a solution to the dilemma, by advocating for the training and inference of recognition models on randomly selected frequency components. We distill our findings into a novel privacy-preserving face recognition method, PartialFace. Extensive experiments demonstrate that PartialFace effectively balances privacy protection goals and recognition accuracy. Code is available at: \url{https://github.com/Tencent/TFace}.

\end{abstract}

%%%%%%%%% BODY TEXT
\section{Introduction}
\label{sec:introduction}

\textit{Face recognition} (FR) is a landmark biometric technique that enables a person to be identified or verified by face. It has seen remarkable methodological breakthroughs and rising adoptions in recent years. Currently, considerable applications of face recognition are carried out online to bypass local resource constraints and to attain high accuracy~\cite{DBLP:journals/air/LiuCWLX23}: Face images are collected by local devices such as cell phones or webcams, then outsourced to a service provider, that uses large convolutional neural networks (CNN) to extract the faces’ identity-representative templates and matches them with records in its database.

During the process, the original face images are often considered sensitive data under regulatory demands that are unwise to share without privacy protection. This fosters the studies of \textit{privacy-preserving face recognition} (PPFR), where cryptographic~\cite{DBLP:conf/icisc/SadeghiSW09,DBLP:conf/pet/ErkinFGKLT09,DBLP:conf/icml/Huang0LA20,DBLP:conf/acmturc/KouZZL21,DBLP:conf/apccas/Ergun14a,DBLP:journals/iotj/MaLLMR19,DBLP:conf/ithings/YangZLLL18,DBLP:journals/soco/XiangTCX16} and perturbation-based~\cite{DBLP:journals/finr/ZhangHXGY20,DBLP:journals/compsec/ChamikaraBKLC20,DBLP:conf/eccv/XuZR18,DBLP:conf/socpar/HondaOUN15,DBLP:conf/btas/MirjaliliRR18,DBLP:journals/ijon/LiWL19b,DBLP:conf/icb/MirjaliliRNR18,DBLP:conf/icb/MirjaliliR17} measures are taken to prevent face images from unauthorized access of, \eg, wiretapping third parties. Face images are converted to protective representations that their visual information is both \textit{concealed} and \textit{cannot be easily recovered}~\cite{electronics9081188}.

This paper advocates a novel PPFR scheme, to learn face images from random combinations of their partial frequency components. Our proposed PartialFace can protect face images' visual information and prevent recovery while maintaining the high distinguishability of their identities.

\begin{figure}[tbp]
  \centering
%   \fbox{\rule{0pt}{2in} \rule{0.9\linewidth}{0pt}}
   \includegraphics[width=\linewidth]{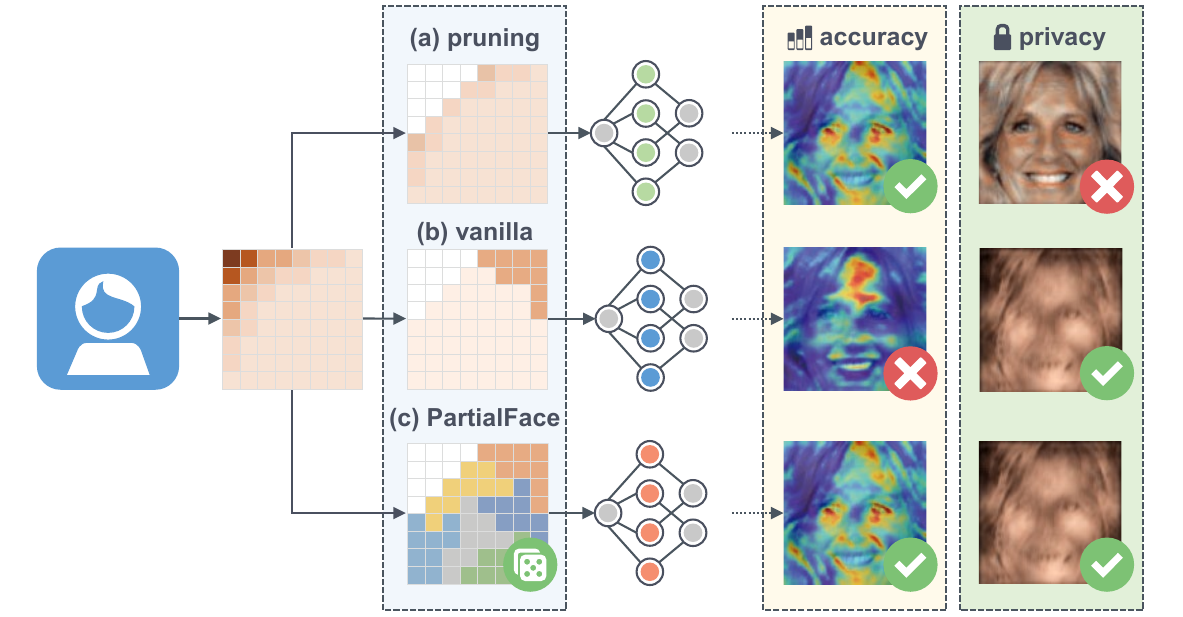}

   \caption{Paradigm comparison among other frequency-based methodologies and PartialFace. (a) Pruning low-frequency channels can conceal visual information but stop no recovery. (b) A vanilla method that uses fixed channel subsets to impede recovery suffers downgraded accuracy. (c) PartialFace addresses the dilemma by training and inferring from random channels.}
   \label{fig:paradigm}
\end{figure}

We start with the disparity in how humans and models perceive images. Recent image classification studies suggest that models’ predictions are determined mainly by the images’ high-frequency components~\cite{DBLP:conf/cvpr/WangWHX20, DBLP:conf/cvpr/0007QSWCR20}, which carry negligible visual information and are barely picked up by humans. We extend the theory to face recognition from a privacy perspective, to train and infer the model on \textit{pruned} frequency inputs: As depicted in~\cref{fig:paradigm}(a), we decouple the face images' frequency compositions via \textit{discrete cosine transform} (DCT), which breakdown every spatial domain into a certain number of (typically 64) constituent bands, \ie, \textit{frequency channels}. We prune the human-perceivable low-frequency channels and exploit the remaining. We find the processed face images become almost visually indiscernible, and the model still works accurately.
% ~\cref{fig:protection}(a) 

Pruning notably conceals visual information. However, to what degree can it impede recovery? We define recovery as the general attempts to reveal the faces' visual appearances from shared protected features using trained attack models. Notice we may prune very few (say, about 10) channels, if according to human perception. At the same time, the remaining high-frequency channels being shared, hence exposed, are quite numerous and carry a wealth of model-perceivable features. While the information abundance can benefit a recognition model, it is also exploitable by a model carrying out attacks, as both may share similar perceptions. Therefore, as we later experimentally show, the attacker can recover visual features from high-frequency channels with ease in the absence of additional safeguards, rendering privacy protection useless.

An intuitive tactic to reinstate protection is to reduce the attacker's exploitable features by training on a small portion of fixed channels, as shown in~\cref{fig:paradigm}(b). However, we find the reduction also severely impairs the accuracy of trained recognition models. 
Evidence on the models' attention, as later shown, attributes their utility downgrade to being incapable of learning a complete set of facial features, as vital channels describing some local features may be pruned.
% ~\cref{fig:observation}

Training on subsets of channels hence seems contradictory to the privacy-accuracy equilibrium. Fortunately, we can offer a reconciliation getting inspired by a recent time-series study~\cite{DBLP:conf/icml/ZhouMWW0022}. It proves under mild conditions, models trained on \textit{random} frequency components can preserve more entirety's information than on \textit{fixed} ones, plausibly by alternately learning from complementary features. We hence propose a novel address to the equilibrium based on its theoretical findings and our observation on model attention: For any incoming face image, we arbitrarily pick a small subset of its high-frequency channels. Therefore, our recognition model is \textit{let trained and inferred from image-wise random chosen channels}, illustrated in~\cref{fig:paradigm}(c).

We further show that randomness can be adjusted to a moderate level, by choosing channels from pre-specified combinations and perturbations called \textit{ranks}, to keep privacy protection while reconciling technical constraints to ease training. At first glance, our randomized approach may seem counter-intuitive as it is common wisdom that models require consistent forms of inputs to learn stably. However, since DCT produces spatially correlated frequency channels that preserve the face's structural information, as later illustrated in~\cref{fig:dct}, it turns out the model generalizes quite naturally. Experimental analyses shows our PartialFace well balances privacy and accuracy. % : The models stop overfitting and recognize accurately while still using significantly fewer channels. 

The contributions of our paper are three-fold:
\begin{enumerate}
    % \item We present an in-depth study of 
    %\item We discover training models on arbitrary ranks can reduce data dimensionality and bring in randomness, to enhance privacy and retain high accuracy.
    % \item We discover that training on random ranks formed by subsets of channels can retain high accuracy while providing enhanced privacy.
    \item We present an in-depth study of the privacy protection of face images, regarding the privacy goals of concealing visual information and impeding recovery.
    \item We propose two methodological advances to fulfill the privacy goals, pruning low-frequency components and using randomly selected channels, based on the observation of model perception and learning behavior.
    \item We distill our findings into a novel PPFR method, PartialFace. We demonstrate by extensive experiments that our proposed method effectively safeguards privacy and maintains satisfactory recognition accuracy.
\end{enumerate}

% \newpage
%-------------------------------------------------------------------------

\begin{figure*}[tbp]
  \centering
%   \fbox{\rule{0pt}{2in} \rule{0.9\linewidth}{0pt}}
   \includegraphics[width=\linewidth]{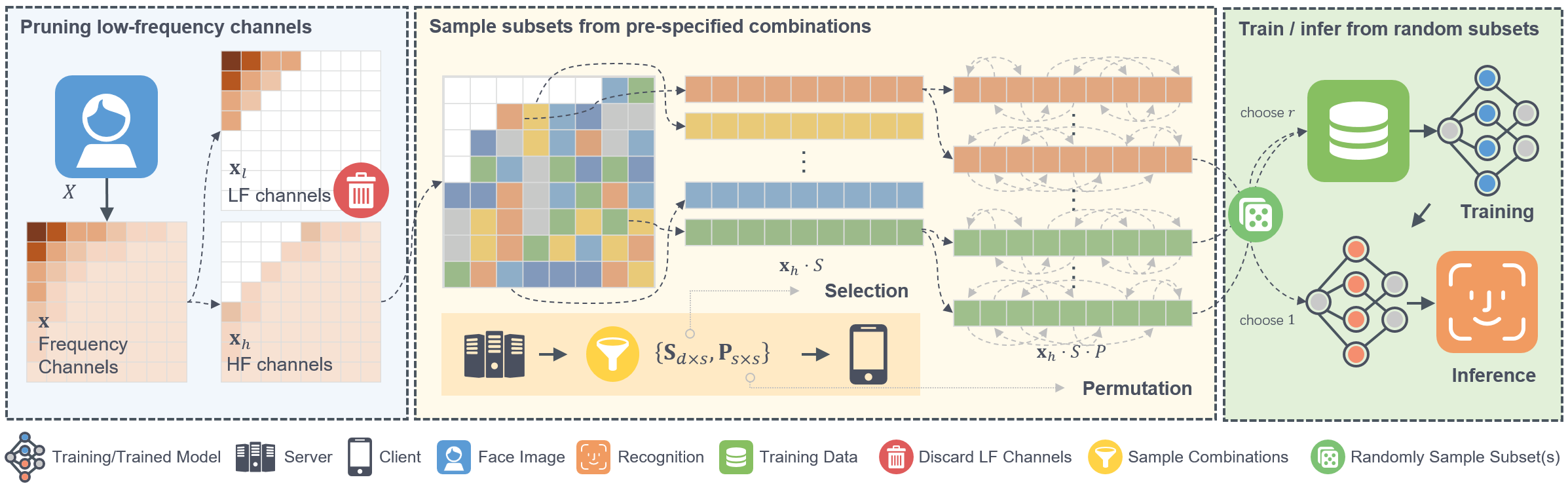}

   \caption{Pipeline of PartialFace. DCT turns face images into the frequency domain, where low-frequency channels are first pruned to remove the human perception. A small subset of channels is selected and permuted at random according to pre-specified combinations $\{\mathbf{S}, \mathbf{P}\}$, or ranks. The model is trained and inferred from random subsets to address the equilibrium between accuracy and privacy.}
   \label{fig:pipeline}
\end{figure*}

\section{Related work}
\label{sec:related-work}

\subsection{Face recognition}
\label{subsec:fr}

The current method of choice for face recognition is CNN-based embedding. The service provider trains a CNN with a softmax-based loss to map face images into one-dimensional embedding features which achieve large inter-identity and small intra-identity discrepancies. While the state-of-the-art (SOTA) FR methods~\cite{DBLP:conf/cvpr/DengGXZ19,DBLP:conf/cvpr/WangWZJGZL018,DBLP:conf/cvpr/LiuWYLRS17} achieve impressive task utility in real-world applications, their attention to privacy protection could be deficient.

\subsection{Privacy-preserving face recognition}

The past decade witnessed significant advances in privacy-preserving face recognition~\cite{DBLP:journals/ijon/WangD21a,DBLP:journals/tifs/MedenRTDKSRPS21,DBLP:journals/corr/abs-2206-10465}. We roughly categorize the related arts into two branches:

\noindent \textbf{Cryptographic methods} perform recognition on encrypted images, or by executing dedicated security protocols. Many pioneering works fall under the category of homomorphic encryption (HE)~\cite{DBLP:conf/icisc/SadeghiSW09,DBLP:conf/pet/ErkinFGKLT09,DBLP:conf/icml/Huang0LA20} or secure multiparty computation (MPC)~\cite{DBLP:journals/iotj/MaLLMR19,DBLP:conf/ithings/YangZLLL18,DBLP:journals/soco/XiangTCX16}, to securely carry out necessary computations such as model's feature extraction. Some methods also employ various crypto-primitives including one-time-pad~\cite{DBLP:conf/apccas/Ergun14a}, matrix encryption~\cite{DBLP:conf/acmturc/KouZZL21}, and functional encryption~\cite{DBLP:conf/pkc/AbdallaBCP15}. The major pain points of these methods are generally high latency and expensive computational costs.

\noindent \textbf{Perturbation-based methods} transform face images into protected representations that are difficult for unauthorized parties to discern or recover. Many methods leverage differential privacy (DP)~\cite{DBLP:journals/finr/ZhangHXGY20,DBLP:journals/compsec/ChamikaraBKLC20,DBLP:journals/ijon/LiWL19b,DBLP:conf/hotedge/X18}, in which face images are obfuscated by a noise mechanism. Some use autoencoders~\cite{DBLP:conf/icb/MirjaliliRNR18} or adversarial generative networks (GAN)~\cite{DBLP:conf/btas/MirjaliliRR18,DBLP:journals/ijon/LiWL19b} to recreate visually distinct faces that maintain a constant identity. Others compress the original images into compact representations by mapping discriminative components into subspaces~\cite{DBLP:conf/autoid/KevenaarSVAZ05,DBLP:conf/mlsp/ChanyaswadCMK16,DBLP:conf/www/MireshghallahTJ21}, or anonymize the images by clustering their features~\cite{DBLP:conf/socpar/HondaOUN15}. These methods face a common bottleneck of task utility: As the techniques they employ essentially distort the original images, their protection often impairs recognition accuracy.

\subsection{Learning in the frequency domain}
Converting spatial or temporal data into the frequency domain provides a powerful way to extract interesting signals. For instance, the Fourier-based \textit{discrete cosine transform}~\cite{DBLP:journals/tcom/ChenSF77} is used by the JPEG standard~\cite{DBLP:journals/cacm/Wallace91} to enable image compression. Researches in deep learning~\cite{DBLP:conf/iclr/GueguenSLY18,DBLP:conf/cvpr/0007QSWCR20} suggest models trained on images' frequency components can perform as well as trained on the original images. Advance~\cite{DBLP:conf/cvpr/WangWHX20} further reveals humans and models perceive low- and high-frequency components differently. In the realm of PPFR, three recent methods~\cite{wang22ppfrfd,DBLP:conf/mm/MiHJLXDZ22,DBLP:conf/eccv/JiWHWXDZCJ22} are closely related to ours as we all conceal visual information by exploiting the split in human and model perceptions. 
However, these methods bear inadequacy in defending recovery, according to our previous discussion of channel redundancy and later testified in experiments.

\section{Methodology}
\label{sec:methods}

This section discusses the motivation and technical details behind PartialFace. PartialFace is named after its key protection mechanism, where the model only exploits face images' partial frequency components to reduce information exposure. \Cref{fig:pipeline} describes its framework.

\subsection{Overview}
\label{subsec:overview}

Recall our privacy goals are to conceal the face images' visual information and to prevent recovery attacks on them. They respectively target the adversarial capability of humans and models. We naturally concretize them into two technical aims, on \textit{how to eliminate human perception}, and, on \textit{how to reduce exploitable information for attack models}.

To eliminate human perception, we leverage the finding that models' utility can be maintained almost in full on the images' high-frequency channels~\cite{DBLP:conf/cvpr/WangWHX20}. Contrarily, humans mostly perceive the images' low-frequency channels, as only they carry signals of conspicuous amplitude for human eyes to discern. Whereby spatial-frequency transforms, these channels can be easily located and pruned from raw images. We concretely opt for DCT as our transform as it facilitates the calculation of \textit{energy}, to serve as the quantification for human perceivable information. Experimentally, we can prune a very small portion of channels to eliminate 95\% of total energy, hence satisfying our aim.

Reducing the attacker's exploitable information requires further pruning of high-frequency channels, as previously discussed. However, it is equally crucial to maintain the recognition model's accuracy, which presents a seeming dilemma, as the training features utilizable by recognition and attack models are tightly interwoven in these channels. We offer a viable way as our major contribution: We notice different channel subsets each carrying certain local facial features. Hence, we feed the model with a random subset from each face image, with the hope to let the model learn the entire face's impression from different images' complementary local feature partitions. This approach is proved feasible by recent theoretical studies~\cite{DBLP:conf/icml/ZhouMWW0022} and demonstrated by our experiments. Therefore, information exposure is minimized as only each image's chosen subset is exposed, and the model still performs surprisingly well.

% We use random channel subsets from each image, that should represent different local facial features, as the model inputs. Hence, it is possible for the model to learning a comprehensive impression of the entire face from these complementary feature partitions.

% and recent theoretical studies~\cite{DBLP:conf/icml/ZhouMWW0022} 
% A viable way offered by our later observation on model attention and recent theoretical studies~\cite{DBLP:conf/icml/ZhouMWW0022}, is to let the model learn complementary channel partitions, each sampled from partial sets of face images. Therefore, only a small channel subset is exposed for each single image, and the model can acquire an understanding of the entirety.

% Fortunately, recent theoretical studies~\cite{DBLP:conf/icml/ZhouMWW0022} along with our observation (elaborated further in~\cref{subsec:rand-recognition}) offer a viable way to preserve recognition performance: We find models are able to learn complementary information from distinct frequency channels of various face images, to acquire an understanding of the entirety. Therefore, we propose to \textit{train and infer models from image-wise random channels}. 

Despite being well justified by theory, we find sampling channels under complete randomness could under-perform in actual training due to two technical limitations: the inadequacy in training samples and the biased sampling within mini-batches. We propose two targeted fixes to reconcile the constraints, by augmenting samples and seeking a moderate level of randomness. We find the modified approach satisfactorily addresses the privacy-accuracy equilibrium.

\subsection{Conceal visual information}
\label{subsec:remove-feature}

\begin{figure}[tbp]
  \centering
%   \fbox{\rule{0pt}{2in} \rule{0.9\linewidth}{0pt}}
   \includegraphics[width=\linewidth]{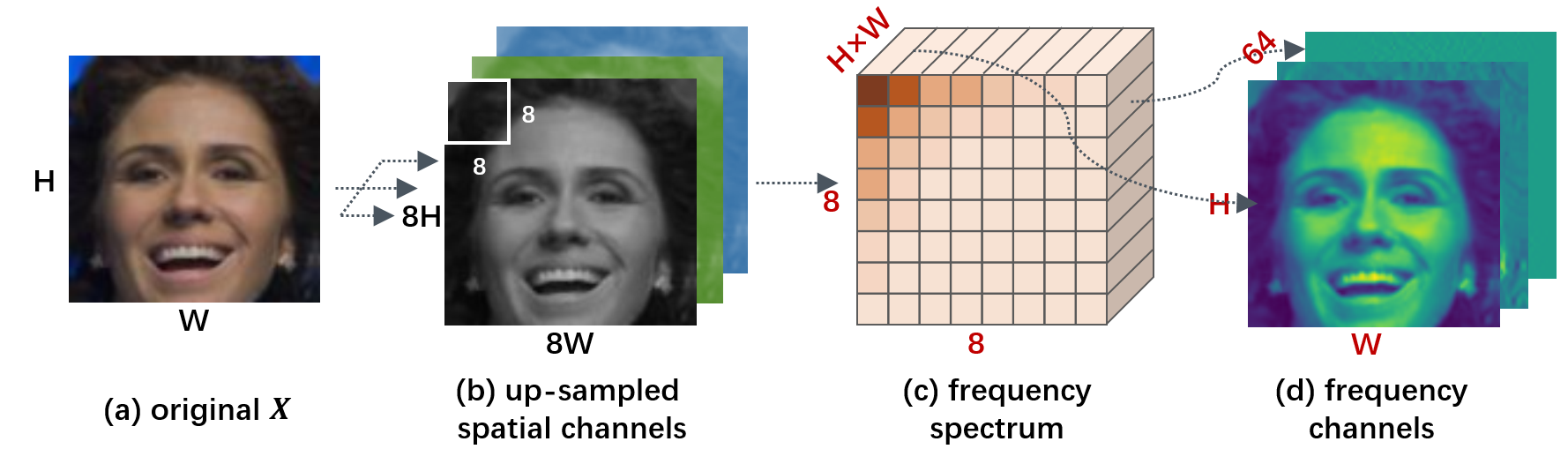}

   \caption{The DCT process. The produced (d) frequency channels keep the spatial structure to (a) the original image, though only low-frequency channels are discernible bu humans. We use (c) 2D grids to describe the frequency spectrum. Each cell in the grid represents one frequency channel of $H$$\times$$W$. }
   \label{fig:dct}
\end{figure}

We first set up some basic notions: $\langle X,y \rangle$ denotes a data sample of a face image and its corresponding label. $\mathbf{x}$ denotes the frequency composition of $X$ and $x_i$ denotes its individual frequency channels. $f(\cdot ; \theta)$ denotes the recognition model parameterized by $\theta$. $l(\cdot, \cdot)$ denotes a generic loss function (\eg, ArcFace). $\mathcal{T}(\cdot)$ denotes the discrete cosine transform and $\mathcal{T}^{-1}(\cdot)$ denotes its inverse transform. % to define, accuracy function, attack model, etc.

To conceal visual information, recall humans and models mainly perceive low- and high-frequency components, respectively. Therefore, we need to find a frequency decomposition of $X$=$\{\mathbf{x}_l,\mathbf{x}_h\}$, where $\mathbf{x}_l,\mathbf{x}_h$ are the respective low- and high-frequency channels, then prune $\mathbf{x}_l$.

We presume $X$ is with the shape of $(H, W)$. We employ DCT to transform $X$'s spatial channels into a frequency spectrum. While an RGB image typically has 3 spatial channels, we pick one for simplicity.  We perform an 8-fold up-sampling ahead to turn $X$ into $(8H, 8W)$. As DCT later divides $H$ and $W$ by 8, this makes sure the resulting frequency channels can be fed into the model as usual. \Cref{fig:dct} illustrates the process of DCT, where $\mathbf{x}$=$\mathcal{T}(X)$.  Concretely, $X$ is divided into $(8,8)$-pixel blocks. DCT turns each block into a 1D array of 64 frequency coefficients and reorganizes all coefficients from the same frequency across blocks into an $(H, W)$ frequency channel (there are 64 of them), that is spatially correlated to the original $X$. As a result, $X$ is turned into $\mathbf{x}$ of $(64, H, W)$.

We then decouple $\mathbf{x}$ into $\{\mathbf{x}_l,\mathbf{x}_h\}$. Notice that human-perceivable low-frequency channels should meanwhile be those with higher amplitude, as humans preferentially identify signals with conspicuous value changes. We measure a channel's amplitude by its \textit{channel energy} $e(\cdot)$, which is the mean of amplitudes of all its elements:

\begin{equation}
    e(x) = \frac{1}{HW} \sum_{i=0}^{H-1}{\sum_{j=0}^{W-1}{|x^{i,j}|}}.
\label{eq:energy}
\end{equation}

We choose $\sigma$=10 highest-energy channels as $\mathbf{x}_l$ to be pruned and the rest as $\mathbf{x}_h$ to be available to models. We experimentally find $\sum_{x\in \mathbf{x}_l}{e(x)} \geq 0.95\sum_{x\in \mathbf{x}}{e(x)}$, and a model trained with $\argmin_\theta{l(f(\mathbf{x}_h,\theta),y)}$ obtain close accuracy to one trained with  $\argmin_\theta{l(f(\mathbf{x},\theta),y)}$. For privacy, example of a channel-pruned image in~\cref{fig:protection}(a) show that most visual information is concealed. While its recovery is still carried out quite successfully, we are to address the issue in the following.

\subsection{Impede easy recovery}
\label{subsec:rand-recognition}

\begin{figure}[tbp]
  \centering
%   \fbox{\rule{0pt}{2in} \rule{0.9\linewidth}{0pt}}
   \includegraphics[width=0.95\linewidth]{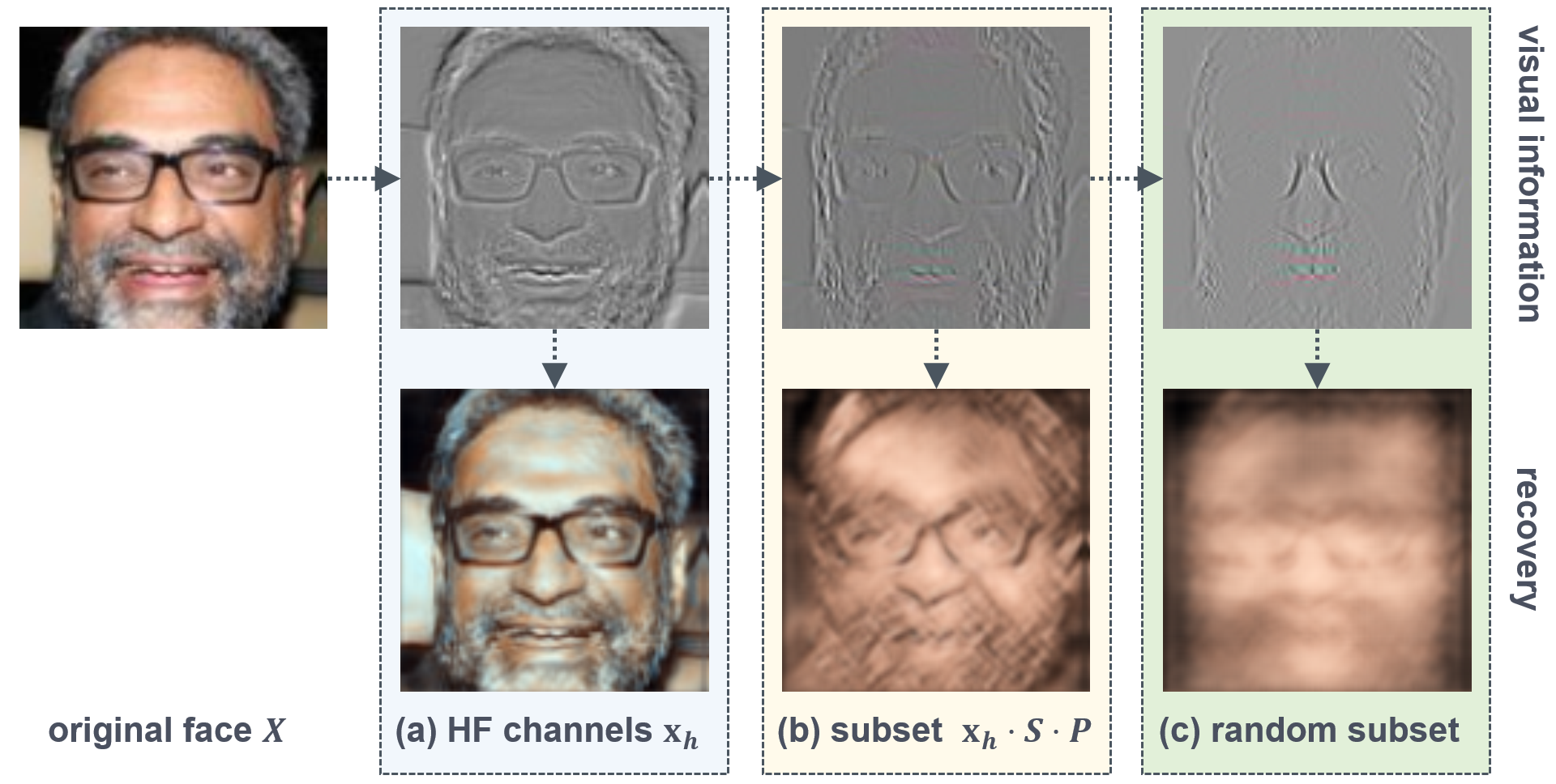}

   \caption{The visual appearance and recovery of an example image of (a) after pruning, (b) a fixed subset, and (c) a random subset. This shows protection is enhanced by removing human perception, reducing the number of channels, and randomness, respectively.}
   \label{fig:protection}
\end{figure}

% , it is imperative to further reduce the model-perceivable channels. However, it is equally crucial to maintain a satisfactory level of recognition performance. This presents a seeming paradox, as the information exploitable by recognition and attack models is tightly interwoven.

To elucidate the seeming dilemma between reducing model-available channels and retaining high recognition accuracy, we first introduce an intuitive protection, referred to as the \textit{vanilla} method. It trains models on a \textit{fixed} small subset of channels straightforwardly:
% To impede recovery, one must first address the seeming paradox between reducing model-exploitable information and retaining high recognition accuracy, as discussed in~\cref{sec:introduction}. We elucidate it with the most intuitive protection measure (here and later called the \textit{vanilla} method), to train models on a \textit{fixed} small subset of channels. 
Concretely for every $X$, the model pick $s$\textless$d$ channels $\mathbf{x}_s$=$(x_{a_1},\dots,x_{a_s})$ from its high-frequency $\mathbf{x}_h$=$(x_1,\dots,x_d)$, where $a_1$\textless$a_2$\textless$\cdots$\textless$a_s$ are fixed indices. We benchmark the trained model on IJB-B/C with $s$=9,18,36 and $d$=54, and verify its privacy. 

\begin{figure}[tbp]
  \centering
%   \fbox{\rule{0pt}{2in} \rule{0.9\linewidth}{0pt}}
   \includegraphics[width=0.95\linewidth]{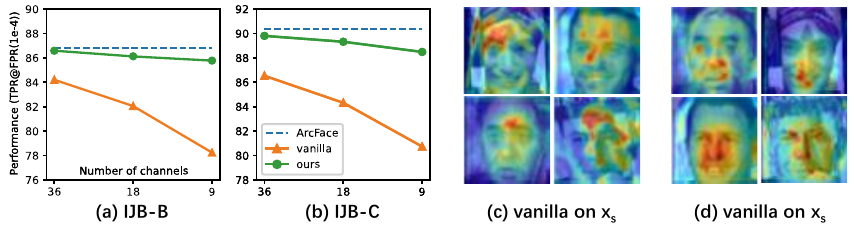}

   \caption{The privacy and accuracy dilemma. (a-b) The vanilla model trained on fixed channels under-performs on IJB-B/C (orange), compared to the unprotected baseline (blue) and PartialFace (green). (c-d) The model gains local attention on foreheads and cheeks, suggesting missing face-describing information. The attention varies by the specific choice of $\mathbf{x}_s$. }
   \label{fig:observation}
\end{figure}

The model indeed effectively prevents recovery, as in~\cref{fig:protection}(b) its recovered image is very blurred. However, the benchmark in~\cref{fig:observation}(a-b) suggests its accuracy dropped by at most 9\%. To find out why is the model's performance impaired, we inspect its attention via Grad-CAM~\cite{DBLP:journals/ijcv/SelvarajuCDVPB20}. Results are exemplified in~\cref{fig:observation}(c-d), which we note show similar patterns among different face images. We derive two observations: (1) Unlike high-utility models that typically have attention to the full face, this model only gains attention to certain local features, which suggests some vital face-describing information is missing in $\mathbf{x}_s$; (2) By training vanilla models on different $\mathbf{x}_s$, we find it can acquire distinct information from different channels, suggesting its attention is correlated to the specific choices of $\mathbf{x}_s$. 

The first observation suggests the cause of the accuracy downgrade. Meanwhile, the second pursues us to consider a viable bypass: We can gather complementary local features from different images and initiate mixed training, to let the model learn a holistic impression of the entirety while keeping the individual information exposure minimized. Our idea is corroborated by~\cite{DBLP:conf/icml/ZhouMWW0022}, which proves in time-series, training models on random frequency components can preserve more information compared to training on fixed ones.

We concrete our theory into a randomized strategy, by advocating training and inferring the recognition model on image-wise randomly chosen channels. Formally, we construct matrix $S \in \{0,1\}^{d\times s}$, with $s_{ij}$=1 if
$i$=$a_j$ and $s_{ij}$=0 if otherwise. We also construct permutation matrix $P \in \{0,1\}^{s\times s}$. For each $X$, we draw $(a_1,\dots,a_s)$ (therefore $S$) and $P$ uniformly at random, and calculate

\begin{equation}
    \mathbf{x}_s = \mathbf{x}_h \cdot S \cdot {P}.
\label{eq:random-sample}
\end{equation}

Here, $S$ randomly pick $s$ channels out of $\mathbf{x}_h$, then their order is permuted by $P$. We introduce $P$ to further impede recovery, as later discussed in~\cref{subsec:enhance-privacy}. The model is trained in the same way as standard FR using $\langle \mathbf{x}_s, y \rangle$. However, it \textit{does not} require to know the sample-wise specific $\{S,P\}$, which is favorable for privacy. The approach sounds a bit counter-intuitive since receiving random inputs seems to mess with the model's understanding. However, recall that the DCT frequency channels are spatially correlated to $X$ and each other. The randomness only pertains to the frequency spectrum while the face's structural information is preserved and unaltered. The model, therefore, can associate different $\mathbf{x}_s$ to the same $X$ quite naturally.

\subsection{Reconcile technical constraints}
\label{subsec:reconcile-constraint}

% Randomness is our core idea in~\cref{subsec:rand-recognition}. However, for models to perform well 

\Cref{eq:random-sample} distills our core idea to sampling $\mathbf{x}_s$ at complete random. However, for practical training, there need to further reconcile two \textit{technical constraints} to achieve satisfying model performance: First to note that $\langle \mathbf{x}_s, y\rangle$ are more varied in feature representations than the unprotected $\langle X, y\rangle$, owing to the randomized frequency. It hence could plausibly take the model with more training samples and a longer training time to approach a well-generalized convergence. Second, ideally, channels at random should be sampled with equal probability to ensure balanced learning of different local features. However, when the training data is partitioned into small mini-batches, sampling is often less unbiased batch-wise so the occurrence of different channel combinations may vary greatly, which we find could undermine the training stability.

Targeting the constraints, we slightly adjust our established approach in two ways. We first augment the training dataset by picking multiple $\mathbf{x}_s$ each time from a face image $X$: From $X$ we sample $\{\mathbf{x}_s^{1},\dots,\mathbf{x}_s^{r}\}$, where $r$ determines the degree of augmentation. Each $\mathbf{x}_s^i$ is independently drawn and appended to the training dataset as individual training samples. The dataset is then fully shuffled so that $\mathbf{x}_s^i$ corresponding to the same face image cannot be easily associated. 

We also control the randomness to a moderate level by specifying the possible combinations of $S,P$ in advance. Concretely, we opt for choosing one $S,P$ from $\mathbf{S}$=$\{S_1,\dots,S_m\}$ and $\mathbf{P}$=$\{P_1,\dots,P_n\}$ respectively, where $\mathbf{S},\mathbf{P}$ are determined by the service provider. We specifically require $\{S_1,\dots, S_n\}$ to be a non-overlapping partition of channels from $\mathbf{x}_h$ (\ie, divide $\mathbf{x}_h$ into equal-length subsets) to maximize the use of $\mathbf{x}_h$ and reduce model bias. Therefore, each $\mathbf{x}_s^i$ is picked from one of $m$$\times$$n$ fixed combinations of channels, called \textit{ranks}. This allows us to facilitate the training and overcome sampling biases. The service provider shares $\{\mathbf{S},\mathbf{P}\}$ with all local devices, so the latter can generate their query $\mathbf{x}_s$ accordingly. During inference, the model is expected to provide \textit{consistent} results of the same query $X$ regardless of the choice of rank, since it learned about a mapping from local features to the face's entirety. We later testify to it in~\cref{subsec:comp-vanilla}. Meanwhile, the model is still unaware of the sample-wise specific choice of $\{S,P\}$ as the recognition relies on nothing else but $\mathbf{x}_s$ alone. Hence privacy is maintained.

To conclude, we present PartialFace that train and infer the recognition model with $\argmin_\theta{l(f(\mathbf{x}_s^i,\theta),y)}$, where $X$=$\{\mathbf{x}_l,\mathbf{x}_h\}$ and $\mathbf{x}_s^i$=$\mathbf{x}_h \cdot S \cdot {P}$ in random, parameterized by $\{\mathbf{S},\mathbf{P}\}$ and $(\sigma,s, r,m,n)$. Later analyses in~\cref{subsec:comp-prior-arts,subsec:comp-vanilla} show PartialFace overcomes the drawback of the vanilla method, plus outperforms most prior arts, to achieve satisfactory recognition accuracy.

\subsection{Enhance privacy with randomness}
\label{subsec:enhance-privacy}

The benefit of randomness is multi-fold. We have discussed it for now on helping address the accuracy and privacy balance and enhance recognition performance. In retrospect, we briefly elaborate on how randomness further safeguards privacy to a large extent.

Recall we remove visual information by pruning $\mathbf{x}_l$ and impede recovery by choosing a subset of $\mathbf{x}_s$. After that, randomness further obstructs recovery: To impose recovery, the attacker exploits not only the channels' information but also their \textit{relative orders and positions} in the frequency spectrum. We introduce $P$ to distort the order of channels to this end. As the recognition model does not require sample-wise $\{S,P\}$, they won't be exposed to the attacker. \Cref{fig:protection}(c) shows the improvement against recovery if the attacker doesn't know the specific choice of subsets.

% The shuffling of training data in~\cref{subsec:reconcile-constraint} is another form of randomness. As each $X$ is turned into $\{\mathbf{x}_s^{1},\dots,\mathbf{x}_s^{r}\}$ during augmentation, there lies a risk that one could piece together them to nullify the protection on $X$. By shuffling the data

% We further explain the shuffling of training data in~\cref{subsec:reconcile-constraint}, which can be regarded as another form of randomness. Recall during augmentation, each $X$ is turned into $\{\mathbf{x}_s^{1},\dots,\mathbf{x}_s^{r}\}$. There is a risk that one could piece together them to nullify the protection on $X$.

% Some related methods~\cite{DBLP:conf/mm/MiHJLXDZ22,DBLP:conf/eccv/JiWHWXDZCJ22} provide limited protection on training phase as they require the original $X$ or most of its frequency components. By randomness, we are allowed to use finely-diced $\mathbf{x}_s$

%-------------------------------------------------------------------------

\section{Experiments}
\label{sec:experiments}

\subsection{Experimental settings}

We compare PartialFace with the unprotected baseline and prior PPFR methods on three criteria: recognition performance, privacy protection of visual information, and that against recovery attack. We further study the computation and cost of PartialFace. We mainly employ an IR-50~\cite{DBLP:conf/cvpr/HeZRS16} trained on the MS1Mv2~\cite{DBLP:conf/eccv/GuoZHHG16} dataset as our FR model, while also using a smaller combination of IR-18 and the BUPT~\cite{DBLP:journals/corr/abs-1911-10692} dataset on some resource-consuming experiments. We set $(\sigma,s,r,m,n)$=$(10,9,18,6,6)$ and use fixed $\{\mathbf{S},\mathbf{P}\}$, if not else specified. Benchmarks are carried out on 5 widely used, regular-size datasets, LFW~\cite{LFWTech}, CFP-FP~\cite{DBLP:conf/wacv/SenguptaCCPCJ16}, AgeDB~\cite{DBLP:conf/cvpr/MoschoglouPSDKZ17}, CPLFW~\cite{CPLFWTech} and CALFW~\cite{DBLP:journals/corr/abs-1708-08197}, and 2 large-scale datasets, IJB-B~\cite{DBLP:conf/cvpr/WhitelamTBMAMKJ17} and IJB-C~\cite{DBLP:conf/icb/MazeADKMO0NACG18}.

\subsection{Benchmarks on recognition accuracy}
\label{subsec:comp-prior-arts}

\noindent \textbf{Compared methods.} We compare PartialFace with 2 unprotected baselines, 4 perturbation-based PPFR methods, and 3 methods base on the frequency domain that share close relation to us. Results are summarized in~\cref{tab:comp-sota}. Here, (1) \textbf{ArcFace}~\cite{DBLP:conf/cvpr/DengGXZ19} denotes the unprotected SOTA trained directly on RGB images; (2) \textbf{ArcFace-FD}~\cite{DBLP:conf/cvpr/0007QSWCR20} is the ArcFace trained on the image’s all frequency channels; (3) \textbf{PEEP}~\cite{DBLP:journals/compsec/ChamikaraBKLC20} is a differential-privacy-based method with a privacy budget $\epsilon$=5; (4) \textbf{Cloak}~\cite{DBLP:conf/www/MireshghallahTJ21} perturbs and compresses its input feature space. Its accuracy-privacy trade-off parameter is set to 100; (5) \textbf{InstaHide}~\cite{DBLP:conf/icml/Huang0LA20} mixes up $k$=2 images and performs a distributed encryption; (6) \textbf{CPGAN}~\cite{DBLP:journals/tifs/TsengW20} generates compressed protected representation by a joint effort of GAN and differential privacy; (7) \textbf{PPFR-FD}\footnote{The results of PPFR-FD on IJB-B is unattainable due its non-disclosure of source code. The rest is quoted from its paper~\cite{wang22ppfrfd}. Please note its experimental condition may have slight inconsistency with ours.}~\cite{wang22ppfrfd} adopts a channel-wise shuffle-and-mix strategy in the frequency domain; (8) \textbf{DCTDP}~\cite{DBLP:conf/eccv/JiWHWXDZCJ22} perturbs the frequency components by a noise disturbance mask with learnable privacy budget, where we set $\epsilon$=1; (9) \textbf{DuetFace}~\cite{DBLP:conf/mm/MiHJLXDZ22} is a two-party framework that employs channel splitting and attention transfer.

\begin{table*}[tbp]
\centering
\begin{tabular}{llccccccc}
\toprule
\textbf{Method}          & \textbf{PPFR} & \multicolumn{1}{l}{\textbf{LFW}} & \multicolumn{1}{l}{\textbf{CFP-FP}} & \multicolumn{1}{l}{\textbf{AgeDB}} & \multicolumn{1}{l}{\textbf{CPLFW}} & \multicolumn{1}{l}{\textbf{CALFW}} & \multicolumn{1}{l}{\textbf{IJB-B}} & \multicolumn{1}{l}{\textbf{IJB-C}} \\ \midrule
ArcFace~\cite{DBLP:conf/cvpr/DengGXZ19}        & No   & 99.77                   & 98.30                      & 97.88                     & 92.77                     & 96.05                     & 94.13                     & 95.60                     \\
ArcFace-FD~\cite{DBLP:conf/cvpr/0007QSWCR20}     & No   & 99.78                   & 98.04                      & 98.10                     & 92.48                     & 96.03                     & 94.08                     & 95.64                     \\
\midrule
PEEP~\cite{DBLP:journals/compsec/ChamikaraBKLC20}            & Yes  & 98.41                   & 74.47                      & 87.47                     & 79.58                     & 90.06                     & 5.82                      & 6.02                      \\
Cloak~\cite{DBLP:conf/www/MireshghallahTJ21}           & Yes  & 98.91                   & 87.97                      & 92.60                     & 83.43                     & 92.18                     & 33.58                     & 33.82                     \\
InstaHide~\cite{DBLP:conf/icml/Huang0LA20}      & Yes  & 96.53                   & 83.20                      & 79.58                     & 81.03                     & 86.24                     & 61.88                     & 69.02                     \\
CPGAN~\cite{DBLP:journals/tifs/TsengW20}          & Yes  & 98.87                   & 94.61                      & 96.98                     & 90.43                     & 94.79                     & 92.67                     & 94.31                         \\
\midrule
PPFR-FD~\cite{wang22ppfrfd}         & Yes  & 99.68                   & 95.04                      & 97.37                     & 90.78                     & 95.72                     & /                   & 94.10                          \\
DCTDP~\cite{DBLP:conf/eccv/JiWHWXDZCJ22}         & Yes  & 99.77                   & 96.97                      & 97.72                     & 91.37                     & 96.05                     & 93.29                    & 94.43                          \\
DuetFace~\cite{DBLP:conf/mm/MiHJLXDZ22} & Yes  & 99.82          & 97.79             & 97.93            & 92.35            & 96.10            &  93.66                 &  95.30                    \\
\textbf{PartialFace (ours)} & Yes & 99.80 &	97.63 & 	97.79 &	92.03 &	96.07 & 93.64 & 94.93 \\ 
\bottomrule
\end{tabular}
\caption{Benchmarks on recognition accuracy. PartialFace is compared with the unprotected baselines and PPFR SOTAs.}
\label{tab:comp-sota}
%\begin{tablenotes}
%\footnotesize
%\item {The results of PPFR-FD on IJB-B is missing due to lack of source code. The rest is quoted from~\cite{wang22ppfrfd}.}
% * The result of PPFR-FD on IJB-C is unavailable due to lack of source code by the time of our submission. The rest is quoted from its paper \cite{wang22ppfrfd}.
%\end{tablenotes}
\end{table*}

\noindent \textbf{Performance.} Results are reported on LFW, CFP-FP, AgeDB, CPLFW, and CALFW by accuracy, and on IJB-B and IJB-C by TPR@FPR(1e-4).  \Cref{tab:comp-sota} shows PartialFace achieves close performance to the unprotected baseline, with a small accuracy gap of $\leq$ 0.8\%. In comparison, perturbation-based methods all generalize unsatisfactorily on large-scale benchmarks. PartialFace outperforms all PPFR prior arts but DuetFace. While DuetFace achieves slightly higher accuracy, its protection is inferior to ours as it only covers the inference phase and can be easily nullified by recovery attacks, later see~\cref{subsec:protect-visual,subsec:protect-against-recon}.

\subsection{Comparison with the vanilla method}
\label{subsec:comp-vanilla}

We elaborate that \textit{randomized} PartialFace outperforms the vanilla method that \textit{fixes} channels in not only recognition accuracy but also robustness. As PartialFace employs data augmentation, for a fair comparison regarding the volume of training data, we consider two experimental settings: the standard PartialFace and one without augmentation ($r$=1). We also set $n$=1 to rule out permutation: $P$ is introduced for privacy and is out of our interest here. There are therefore $m$$\times$$n$=6 combinations of channels, or ranks. As channels from different ranks carry distinct information that may affect the performance, we train one vanilla model on each rank and average their test accuracy, with the range marked in parentheses. Though PartialFace is able to infer from arbitrary rank, we evaluate its robustness by inferring from each fixed rank separately. The model is robust if it performs on different ranks consistently (small range). Results are reported on IJB-B/C by TPR@FPR(1e-4) in~\cref{tab:comp-vanilla}.

\begin{table}[tbp]
\centering
\begin{tabular}{lll}
\toprule
\textbf{Method} & \textbf{IJB-B \small{(range)}} & \textbf{IJB-C \small{(range)}} \\
\midrule
ArcFace         & 86.83                  & 90.35                  \\
Vanilla         & 78.24 \small{(-14.49/7.36)}   & 80.73 \small{(-16.37/7.45)}    \\
PF w/o aug.    & 81.43 \small{(-4.19/2.88)}    & 82.81 \small{(-4.29/2.79)}    \\
\textbf{PartialFace}       & 85.77 \small{(-1.95/1.25)}    & 88.48 \small{(-1.71/1.18)}    \\
\bottomrule
\end{tabular}
\caption{Comparison with the vanilla method by accuracy and robustness. Experiments are conducted on IR-18 + BUPT. ``PF w/o aug.'' indicates PartialFace without augmentation.}
\label{tab:comp-vanilla}
\end{table}

\noindent \textbf{Performance.} PartialFace achieves an average accuracy gain of 7.53\% and 7.75\% on IJB-B/C, respectively, compared to the vanilla. The performance is close to the unprotected baseline, showing that the privacy-accuracy trade-off of PartialFace is highly efficient. Even PartialFace without augmentation outperforms the vanilla for about 3\%. We further note: (1) The range shows PartialFace is more robust than the vanilla under different $S_i$. (2) PartialFace outperforms the vanilla for all individual $S_i$. This suggests that randomness empowers PartialFace with the knowledge of the entirety instead of that of certain informative ranks.

\begin{figure}[tbp]
  \centering
%   \fbox{\rule{0pt}{2in} \rule{0.9\linewidth}{0pt}}
   \includegraphics[width=0.95\linewidth]{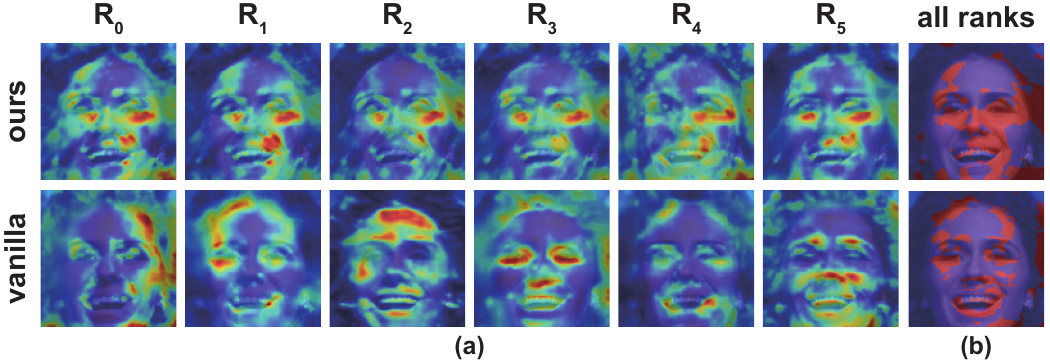}

   \caption{Visualization of model attention via Grad-CAM. (a) PartialFace (1st row) compared with vanilla models (2nd row) on each of 6 different ranks. (b) Integrated attention of all ranks.}
   \label{fig:visual-gradcam}
\end{figure}

\noindent \textbf{Visualization.} We visualize the rank-wise attention of PartialFace and the vanilla via Grad-CAM~\cite{DBLP:journals/ijcv/SelvarajuCDVPB20}, see~\cref{fig:visual-gradcam}(a). The attention of each vanilla model is restrained to local facial features, which indicates the inadequate learning of the entirety features. PartialFace generate accurate attention on the entire face regardless the rank. We integrated the attention across all ranks in~\cref{fig:visual-gradcam}(b). All vanilla models' attention combined gains attention on the entire face, which testifies to our ``learn-entirety-from-local'' theory. Also to note that PartialFace has similar attention across all ranks. This allows it to recognize $X$ using arbitrary rank, as they produce aligned outcomes.

\subsection{Protection of visual information}
\label{subsec:protect-visual}

We investigate PartialFace's protection of privacy. We reiterate that the first privacy goal is to conceal the visual information of face images. We compare ParitalFace with 3 PPFR SOTAs using the frequency domain: PPFR-FD~\cite{wang22ppfrfd}, DCTDP~\cite{DBLP:conf/eccv/JiWHWXDZCJ22}, and DuetFace~\cite{DBLP:conf/mm/MiHJLXDZ22}. These prior arts are closely related to ours since we all leverage the perceptual difference between humans and models as means of protection. However, they differ in the processing of frequency components: Both PPFR-FD and DCTDP remove the component with the highest energy (the DC component). To obfuscate the remaining components, PPFR-FD employs mix-and-shuffle and DCTDP applies a noise mechanism. DuetFace is the most related method in the pruning of low-frequency components. We note in special that they \textit{retain 36, 63, and 54 high-frequency channels}, respectively, while \textit{PartialFace retains 9}. We now demonstrate the methodological differences and varied number of channels result in contrasting privacy protection capacities.

\begin{figure}[tbp]
  \centering
%   \fbox{\rule{0pt}{2in} \rule{0.9\linewidth}{0pt}}
   \includegraphics[width=0.95\linewidth]{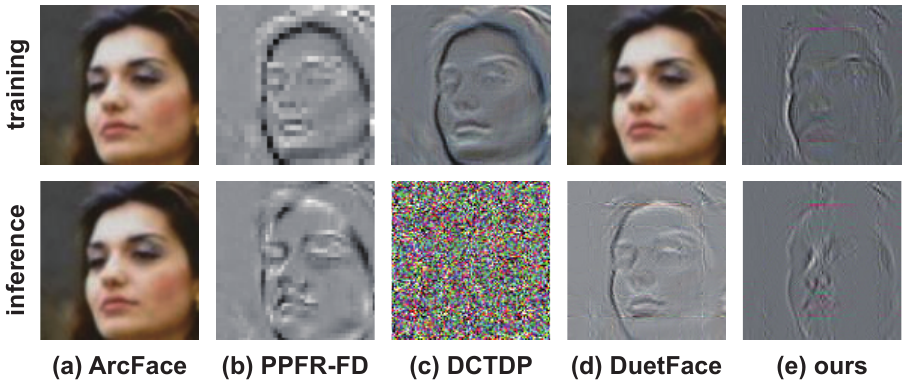}

   \caption{Example face images of (a) unprotected ArcFace, (b-d) compared SOTAs, and (e) PartialFace, during both training and inference phase. PartialFace best conceals visual information.}
   \label{fig:visual-feature}
\end{figure}

\Cref{fig:visual-feature} exemplifies face images processed (therefore are supposed to be protective) by each compared method. As all methods carry out protection in the frequency domain, we convert images back by padding any removed frequency components with zero and applying an inverse transform. PartialFace, \eg, has $X'$=$\mathcal{T}^{-1}(\{\mathbf{0},\mathbf{x}_h\})$.
We visualize the training and inference phases separately, as the compared methods' processing on them may vary: During inference, we argue PPFR-FD and DuetFace (\cref{fig:visual-feature}(b)(d)) provide inadequate protection, as one can still discern the face quite clearly in their processed images. DCTDP effectively conceals visual information after obfuscating the channels with its proposed noise mask (\cref{fig:visual-feature}(c)). However, its protection during training is impaired, as it requires access to abundant non-obfuscated components to learn the mask. DuetFace manifests a similar weakness as it relies on original images to rectify the model's attention (\cref{fig:visual-feature}(d)). 

PartialFace first discards most visual information by pruning $\mathbf{x}_l$. The rest is further partitioned when sampling $\mathbf{x}_s$ from $\mathbf{x}_h$. By energy, each $\mathbf{x}_s$ carries less than 1\% of visual information compared to the original $X$. As a result,~\cref{fig:visual-feature}(e) shows PartialFace provides better protection on visual information than related SOTAs in both phases.

\subsection{Protection against recovery}
\label{subsec:protect-against-recon}

We here provide an in-depth analysis of how PartialFace impedes recovery. Our primary aiming is to show how the removal of model-exploitable information and the randomness of channels contribute to the defense. Assume the recovery attacker possesses a collection of face images $\{X\}$ and is aware of the protection mechanism. It can locally generate protected representations $X'$ ($\mathbf{x}_h$ in our case) from $X$, then train a malicious model $g(\cdot)$ with $\argmin_{\delta}l'(g(X',\delta),X)$, with the aiming to inversely fit $X$ from $X'$. Having intercepted a recognition query, the attacker leverages $g(\cdot)$ to recover the concealed visual information. Upon the knowledge it possesses, a black-box attacker is one unknowing of necessary parameters ($\{\mathbf{S},\mathbf{P}\}$ in our case), whilst a white-box attacker possesses such knowledge. We also study the capability of PPFR-FD, DCTDP, and DuetFace against the white-box attacker. We further investigate an enhanced white-box attacker, who imposes threats dedicated to the randomness of PartialFace.

\begin{figure}[tbp]
  \centering
%   \fbox{\rule{0pt}{2in} \rule{0.9\linewidth}{0pt}}
   \includegraphics[width=0.95\linewidth]{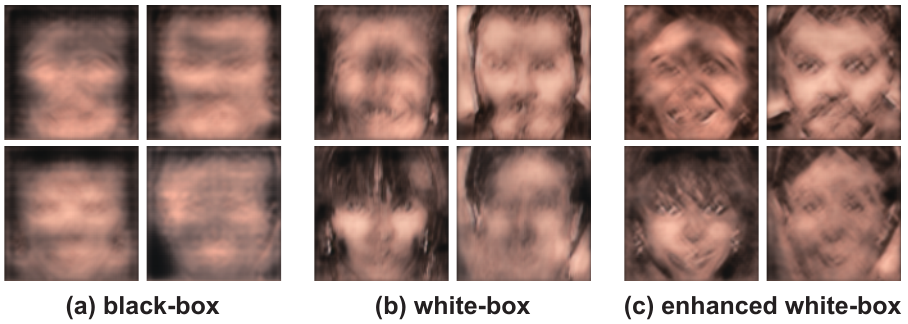}

   \caption{Examples of recovered face images of PartialFace under attackers with varied capabilities: (a) a black-box, (b) a white-box, and (c) an enhanced white-box attacker. PartialFace effectively prevents recovery as all images are blurry and hardly identifiable.}
   \label{fig:recon-b}
\end{figure}

 % We train each model for 24 epochs. Note that the adversarial setting is stronger than in prior works in data volume and model capacity.

\noindent \textbf{Black-box attacker.} It is concretized as a malicious third party, uninvolved in the recognition yet eager to wiretap the transmission. For each attacker, We employ a full-scale U-Net~\cite{DBLP:conf/miccai/RonnebergerFB15} as the recovery model and train it on BUPT. Unknowing of $\{\mathbf{S},\mathbf{P}\}$, it must produce $X'$ based on its own conjectured $\{\mathbf{S}',\mathbf{P}'\}$, which is believably inconsistent with that applied to the recognition model. Thus produced $X'$ are invalid samples and training $g(\cdot)$ on them will nullify the recovery, as shown in~\cref{fig:recon-b}(a).

\noindent \textbf{White-box attacker.} The candidate combinations $\{\mathbf{S},\mathbf{P}\}$ is known by any party participates in the recognition, \eg, an honest-but-curious server. Such an attacker can generate $X'$ correctly. However, receiving a query $\mathbf{x}_s$, the attacker is unknowing of its channels' position and order, since the \textit{specific} $\{S,P\}$ doesn't come with $\mathbf{x}_s$. The missing information is vital to recovery. Consequently, the attack is obstructed by $\mathbf{x}_s$'s randomness in~\cref{fig:recon-b}(b): The recovered image is blurry and hardly identifiable.

\begin{figure}[tbp]
  \centering
%   \fbox{\rule{0pt}{2in} \rule{0.9\linewidth}{0pt}}
   \includegraphics[width=0.85\linewidth]{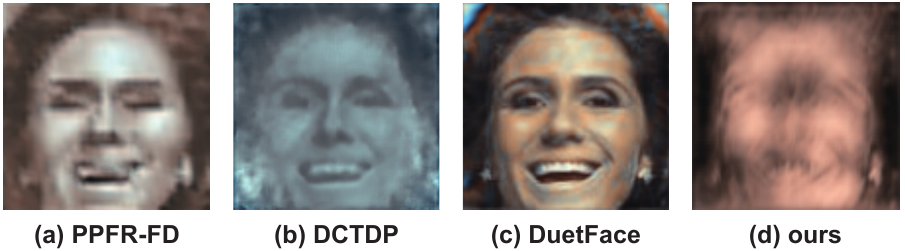}

   \caption{Recovered images from (a-c) compared SOTAs and (d) PartialFace. PartialFace impedes recovery better than the rest.}
   \label{fig:recon-a}
\end{figure}

We plus compare with PPFR-FD, DCTDP and DuetFace under the white-box settings. Among them, DuetFace shows almost no resistance to recovery (\cref{fig:recon-a}(c)). PPFR-FD and DCTDP provide inadequate protection, as images recovered from them (\cref{fig:recon-a}(a-b)) are blurred yet still clearly identifiable. The protection is impaired as their processed images retain most high-frequency components, and the excessive perceivable information is learned by the recovery model. In comparison, PartialFace (\cref{fig:recon-a}(d)) offers significantly outperformed protection.

\noindent \textbf{Enhanced white-box attacker.} A resource-unbounded attacker may brute-forcibly break the randomness of PartialFace leveraging a more sophisticated attack: it trains a series of attack models, one for each candidate $\{S, P\}$. Receiving $\mathbf{x}_s$, the attacker feeds it into every model to try every combination of $\{S, P\}$, until one produces the best recovery. The vague face images in~\cref{fig:recon-b}(c) imply the attempted recovery is unsuccessful, even if the attacker finds the correct $\{S, P\}$. This is attributed to the reduction of model-exploitable channels.

\begin{table}[tbp]
\begin{tabular}{lccc}
\toprule
\textbf{Method}              & \textbf{SSIM \small{(↓)}}  & \textbf{PSNR \small{(↓)}} & \textbf{Accuracy \small{(↓)}} \\
\midrule
PPFR-FD           & 0.713  & 15.66 & 83.73     \\
DCTDP           & 0.687 & 15.42  & 79.60      \\
DuetFace           & 0.866 & 19.88 & 96.52        \\
\textbf{PartialFace}           & 0.591 & 13.70 & 65.35       \\
\bottomrule
\end{tabular}
\caption{Quantitative analyses of the recovery quality. Lower SSIM, PSNR and accuracy suggest better protection. Here, we remind the accuracy of verification is lower bounded by 50.00. }
\label{tab:quant-recon}
\end{table}

\noindent \textbf{Quantitative comparison.} To quantitatively assess the recovery quality, we measure the average structural similarity index (SSIM) and peak signal-to-noise ratio (PSNR) of the recovered images. Additionally, to study if the attacker can exploit its outcome for recognition proposes, we feed the images into a pre-trained model and measure verification accuracy. Lower SSIM, PSNR, and recognition accuracy suggest deficient recovery, thus indicating a higher level of protection. The results in~\cref{tab:quant-recon} suggest PartialFace outperforms its competitors in all three evaluated metrics.

\subsection{Protection against recovery from templates}
\label{subsec:recon-template}

\begin{figure}[tbp]
  \centering
%   \fbox{\rule{0pt}{2in} \rule{0.9\linewidth}{0pt}}
   \includegraphics[width=0.85\linewidth]{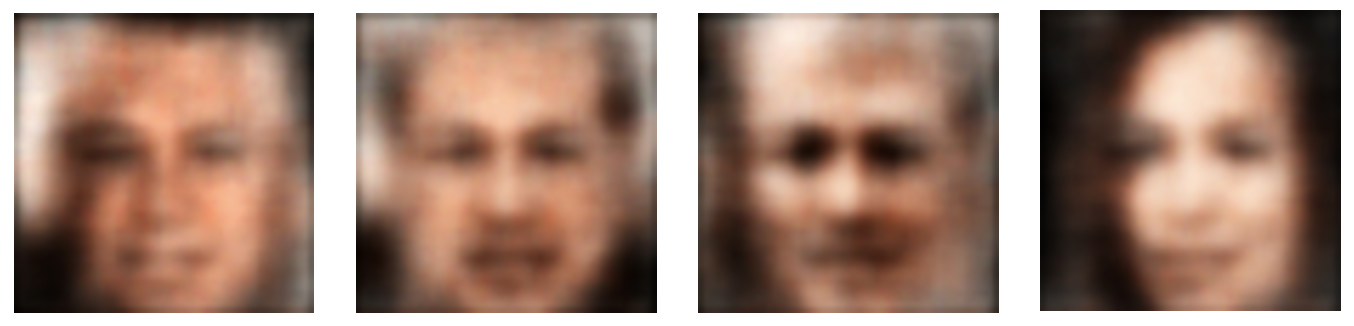}

   \caption{Recovered images from extracted identity templates, that are blurry and can hardly be inferred by humans or models.}
   \label{fig:recon-template}
\end{figure}

By the principle of FR, all $\mathbf{x}_s$ of face image $X$ are later extracted to very similar identity templates, so the recognition on them produces aligned results. Here, one would arise concern that recovery can also be carried out on the extracted templates, which theoretically contain the faces' full identity information, to realize better inversion of images. We demonstrate that such a proposed attack is also ineffective to PartialFace.

Although there could be dedicated attacks on templates, in practice, high-quality recovery from them is difficult as templates are far more compact representations than images and channel subsets. In~\cref{fig:recon-template}, the recovered images are highly blurred, and can hardly be inferred by humans or models, hence won't impose an effective threat.

\subsection{Ablation study}
\label{subsec:ablation}

PartialFace is parameterized by $(\sigma,s,r,m,n)$. Among them, we note $\sigma$ is chosen according to channel energy, and $s$ is determined by $(\sigma, m)$. \Cref{tab:abal} analyzes the choice of the remaining parameters. Results show augmentation ($r$) enhances the model's performance, at the cost of taking longer time to train. $m$ affects the performance mainly by its influence on $s$, as larger $m$ leading to fewer channels in each $\mathbf{x}_s$. We introduce $P$ solely for privacy purposes. Results on $n$ indicate a trade-off between accuracy and privacy. Generally, PartialFace is robust to the choice of parameters.

\begin{table}[tbp]
\centering
\begin{tabular}{lcp{9mm}p{9mm}p{9mm}p{9mm}p{9mm}}
\toprule
\multicolumn{2}{c}{\textbf{Settings}}       & \multicolumn{5}{c}{\textbf{LFW~~~CFP-FP~~AgeDB CPLFW CALFW}} \\ 
\midrule
%\multicolumn{2}{c}{\textbf{Settings}}       & \small{\textbf{LFW}} & \small{\textbf{CFP-FP}} & \small{\textbf{AGEDB}} & \small{\textbf{CPLFW}} & \small{\textbf{CALFW}} \\
\multicolumn{2}{l}{ArcFace}                 & 99.38        & 92.31           & 94.65          & 89.41          & 94.78          \\
\midrule
\multicolumn{1}{c}{\multirow{2}{*}{$r$}} & 36  & 99.51        & 91.27           & 94.44          & 88.93          & 94.81          \\
\multicolumn{1}{c}{}                   & 6  & 98.69        & 86.79           & 90.65          & 84.50          & 92.67          \\
\midrule
\multicolumn{1}{c}{\multirow{2}{*}{$m$}} & 3  & 99.42        & 91.68           & 93.95          & 88.70          & 94.69          \\
\multicolumn{1}{c}{}                   & 9  & 99.32        & 90.74           & 93.56          & 87.30          & 94.36          \\
\midrule
\multirow{2}{*}{$n$}                     & 1  & 99.37        & 91.55           & 94.25          & 88.73          & 94.70          \\
                                       & 12 & 99.35        & 89.21           & 93.17          & 87.19          & 94.26          \\
\midrule
\multicolumn{2}{l}{\textbf{Default}}                 & 99.38        & 91.20           & 93.72          & 88.11          & 94.42         \\
\bottomrule
\end{tabular}
\caption{Ablation study on combinations of hyperparameters. Experiments are carried out on IR-18 + BUPT. When changing one of them, we keep the rest to default values, \ie, $(r,m,n)$=$(18,6,6)$.}
\label{tab:abal}
\end{table}

\subsection{Complexity and compatibility}
\label{subsec:complexity-cost}

\begin{table}[tbp]
\centering
\begin{tabular}{lcccc}
\toprule
\textbf{Settings}       & \textbf{LFW} & \textbf{CFP-FP} & \textbf{AgeDB} & \textbf{CPLFW} \\ 
\midrule
%\multicolumn{2}{c}{\textbf{Settings}}       & \small{\textbf{LFW}} & \small{\textbf{CFP-FP}} & \small{\textbf{AGEDB}} & \small{\textbf{CPLFW}} & \small{\textbf{CALFW}} \\
CosFace                 & 99.53 & 92.89 & 95.15 & 89.52 \\
\textbf{PartialFace}                 & 99.35 & 89.54 & 93.97 & 87.90          \\
\bottomrule
\end{tabular}
\caption{Compatibility of PartialFace. FR models are trained using CosFace on IR-18 + BUPT. Combining PartialFace with CosFace also demonstrates high utility, compared to its baseline.}
\label{tab:compatibility}
\end{table}

Though PartialFace is proposed by studying the model's behavior, privacy protection is solely realized by processing the face images. The decoupling with model architecture and training tactics benefits resources and compatibility. 

\noindent \textbf{PartialFace is resource-efficient.} Compared to the unprotected baseline, PartialFace doesn't increment model size as they share identical model architectures. Training does take more ($r$) time and storage due to augmentation, while we argue it is acceptable for the service provider. The crucial inference time remains the same as the baseline, as DCT and sampling $\mathbf{x}_s$ only increase negligible computation costs.

\noindent \textbf{PartialFace is well compatible.} The decoupling of preprocessing and training also allows PartialFace to serve as a convenient plug-in: PartialFace can be integrated with SOTA FR methods to enjoy enhanced privacy protection. Specifically, we demonstrate the recognition accuracy on CosFace~\cite{DBLP:conf/cvpr/WangWZJGZL018}, a major competitor of ArcFace, in~\cref{tab:compatibility}. Combining PartialFace with CosFace also results in high utility, as compared to its baseline.

\section{Conclusion}

This paper presents an in-depth study of the privacy protection of face images. Based on the observations on model perception and training behavior, we present two methodological advances, pruning low-frequency components and using randomly selected channels, to address the privacy goal of concealing visual information and impeding recovery. We distill our findings into a novel privacy-preserving face recognition method, PartialFace. Extensive experiments demonstrate that PartialFace effectively balances privacy protection goals and recognition accuracy.

% The ubiquitous use of face recognition has sparked increasing privacy concerns, as unauthorized access to sensitive face images could compromise the information of individuals. This paper presents an in-depth study of the privacy protection of face images' visual information and against recovery. Drawing on the perceptual disparity between humans and models, we propose to conceal visual information by pruning human-perceivable low-frequency components. For impeding recovery, we first elucidate the seeming paradox between reducing model-exploitable information and retaining high recognition accuracy. Based on recent theoretical insights and our observation on model attention, we propose a solution to the dilemma, by advocating for the training and inference of recognition models on randomly selected frequency components. We distill our findings into a novel privacy-preserving face recognition method, PartialFace. Extensive experiments demonstrate that PartialFace effectively balances privacy protection goals and recognition accuracy.

%%%%%%%%% REFERENCES
{\small
\bibliographystyle{ieee_fullname}
\bibliography{PartialFace}
}

\end{document}